\documentclass[conference, a4apers]{IEEEtran}
\usepackage{cite}
\usepackage{amsmath,amssymb,amsfonts}
\usepackage{algorithmic}
\usepackage{graphicx}
\usepackage{textcomp}
\usepackage{xcolor}
\usepackage{multirow}

\usepackage[linesnumbered,ruled,vlined]{algorithm2e}
\SetKwInput{KwInput}{Input}                

\begin{document}
%

\title{A Novel Self-Knowledge Distillation Approach with \\ Siamese Representation Learning for \\ Action Recognition}


\author{\IEEEauthorblockN{Duc-Quang Vu}
\IEEEauthorblockA{Dept. of CSIE\\
National Central University\\
Taoyuan, Taiwan\\
Email: vdquang@g.ncu.edu.tw}
\and
\IEEEauthorblockN{Thi-Thu-Trang Phung}
\IEEEauthorblockA{Thai Nguyen University\\
Thai Nguyen, Vietnam\\
Email: phungthutrang.sfl@tnu.edu.vn}
\and
\IEEEauthorblockN{Jia-Ching Wang}
\IEEEauthorblockA{Dept. of CSIE\\
National Central University\\
Taoyuan, Taiwan\\
Email: jcw@csie.ncu.edu.tw}}


\maketitle

\begin{abstract}
Knowledge distillation is an effective transfer of knowledge from a heavy network (teacher) to a small network (student) to boost students' performance. Self-knowledge distillation, the special case of knowledge distillation, has been proposed to remove the large teacher network training process while preserving the student's performance. This paper introduces a novel Self-knowledge distillation approach via Siamese representation learning, which minimizes the difference between two representation vectors of the two different views from a given sample. Our proposed method, SKD-SRL, utilizes both soft label distillation and the similarity of representation vectors. Therefore, SKD-SRL can generate more consistent predictions and representations in various views of the same data point. Our benchmark has been evaluated on various standard datasets. The experimental results have shown that SKD-SRL significantly improves the accuracy compared to existing supervised learning and knowledge distillation methods regardless of the networks.

\end{abstract}

\IEEEpeerreviewmaketitle

\section{Introduction}
\label{sec:introduction}

Action recognition is one of the most important issues in computer vision. Various methods have been proposed to address this task, such as HOG3D~\cite{klaser2008spatio}, SIFT3D~\cite{scovanner20073}, ESURF~\cite{willems2008efficient}, MBH~\cite{dalal2006human}, iDTs~\cite{wang2013action}, etc. Instead of using the traditional methods above to extract features, deep learning models are now trained to automatically learn the features by using convolutional neural networks (CNNs)~\cite{goodfellow2016deep}, and they have brought a large change for computer vision and image processing. Various CNN models have been proposed to address the action recognition task in recent years, such as I3D~\cite{i3d_2017}, SlowFastNet~\cite{SlowFast}, 3D ResNet~\cite{resnet3D_50}, ip-CSN~\cite{tran2019video}, ir-CSN~\cite{tran2019video}, etc. However, these deep models require a lot of layers with millions of parameters. Thus, such approaches may not be suitable for deploying these deep models on embedded or mobile devices with limited resources.

\begin{figure}[!ht]
  \centering
  \centerline{\includegraphics[scale=0.42]{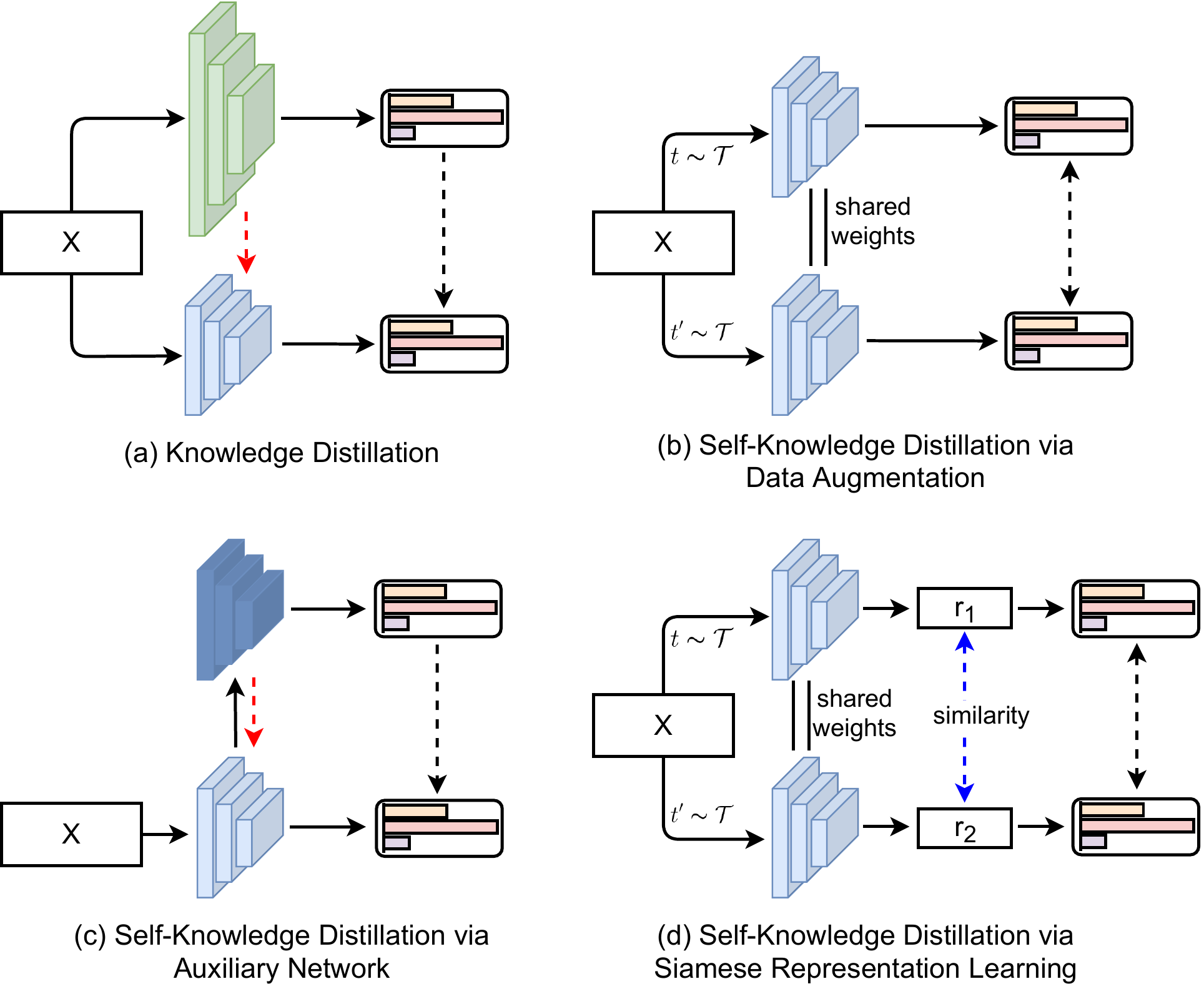}}
  \caption{\textbf{Comparison of various distillation approaches}. The black line is the forward path; the black dashed line indicates the soft label distillation; the red dashed line denotes the feature distillation at intermediate layers; the blue dashed line illustrates the similarity loss between two representation vectors of the same sample. Two data augmentation subsets are sampled from the set of augmentations ($t \thicksim \mathcal{T}$ and $t' \thicksim \mathcal{T}$) and applied to each data to obtain two different views. (a) Conventional knowledge distillation with the heavy teacher (the green network). (b) Self-Knowledge distillation via data augmentation. (c) Self-Knowledge distillation by an auxiliary network (the dark blue network). (d) Our proposed approach with Siamese representation learning.}
  \medskip
  \label{fig:overview}
\end{figure}

Knowledge distillation (KD) has become a promising approach to address the above limitation via transferring knowledge from a larger deep neural network (i.e., the teacher network) to a small network (i.e., the student network). Various KD approaches ~\cite{Hinton_2015_NIPS, Adriana_2015_Fitnets, t3d, crasto2019mars, diba2018spatio} have been proposed to transfer different knowledge, such as using logit vectors from the last layer or feature maps from intermediate layers of a large network to guide the learning of the student model (see Fig.~\ref{fig:overview} (a)). Although this approach can effectively improve the performance of the student network, the existence of the teacher network complicates the training process of a single network, especially for 3D CNNs, when the costs of the training time and the GPU space spend a lot much compared to 2D CNNs. Self-knowledge distillation (Self-KD) has been proposed to remove the large and expensive teacher network. In Self-KD, the student is learned and distilled the knowledge by itself without using any teacher. Various Self-KD-based methods have been proposed for many tasks such as image classification~\cite{yun2020regularizing, ji2021refine}, object detection~\cite{kim2020self}, machine translation~\cite{kim2020self}, natural language processing~\cite{hahn2019self}, etc. In~\cite{xu2019data}, Xu et al. have demonstrated that the Self-KD approach outperforms conventional KD-based methods without training teacher networks.

Self-KD has largely been divided into two main categories: data augmentation-based approaches and auxiliary network-based approaches. Data augmentation-based approaches usually generate a consistent prediction between two different distorted versions of a sample or a pair~\cite{xu2019data} of samples of the same class~\cite{yun2020regularizing} (see Fig.~\ref{fig:overview} (b)). Meanwhile, the auxiliary network-based approaches utilize additional layers in the middle of the classifier network to transfer the knowledge for itself via feature maps at intermediate layers and/or soft label distillation at the classification layer~\cite{ji2021refine} (see Fig.~\ref{fig:overview} (c)). 

In this paper, we introduce a novel self-knowledge distillation approach with Siamese representation learning (SKD-SRL) for action recognition. Siamese networks are weight-sharing neural networks applied on two or more inputs~\cite{chen2020exploring}, and they are usually used to maximize the similarity object in different conditions. Various Siamese network-based methods have been proposed and achieve state-of-the-art performance in self-supervised learning such as SimCLR~\cite{chen2020simple}, SwAV~\cite{caron2020unsupervised}, Barlow Twins~\cite{zbontar2021barlow}, etc. Siamese representation learning is a simple yet robust method to maximizes the similarity of two views from a given input. They are usually applied in unsupervised visual representation learning~\cite{chen2020exploring}. In SKD-SRL, we focus on both the consistency between the two predictive distributions (via soft label distillation) and the similarity between two representation vectors of two distorted views (by Siamese representation learning) in the same sample (see Fig.~\ref{fig:overview} (d)). Experimental results show that our proposed SKD-SRL has significantly improved the accuracy compared to the independently training and conventional KD methods. Moreover, our method outperforms the state-of-the-art supervised learning and KD methods. The details are discussed in Section~\ref{sec:experiment}.

\section{Related Work}
\label{sec:related_work}
Many different approaches and network architectures have been proposed for action recognition in recent years. FASTER~\cite{zhu2020faster} is proposed to learn the predictions from models of different complexities to capture subtle motion information and lightweight representations from cheap models to cover scene changes in the video. Combine with a recurrent network, the FASTER significantly reduces computational cost while maintaining state-of-the-art accuracy across popular datasets. In~\cite{NEURIPS2019_3d779cae}, the authors presented a new architecture based on a combination of a deep subnet operating on low-resolution frames with a compact subnet operating on high-resolution frames. MoViNets~\cite{kondratyuk2021movinets} is introduced as a family of computation and memory-efficient video networks found by neural architecture search. These methods are demonstrated that they require less computational cost and memory while achieving state-of-the-art performance. For distillation approaches, Crasto et al.~\cite{crasto2019mars} proposed a new approach, namely MARS, that allows the knowledge distillation from a flow network, i.e., teacher, to an RGB network, i.e., student. Besides, several approaches based on transformer have been proposed and achieved state-of-the-art performance, such as MViT~\cite{fan2021multiscale}, VTN~\cite{neimark2021video}, etc.

\section{Proposed Method}
\label{sec:method}

\begin{figure}[!ht]
  \centering
  \centerline{\includegraphics[scale=0.4]{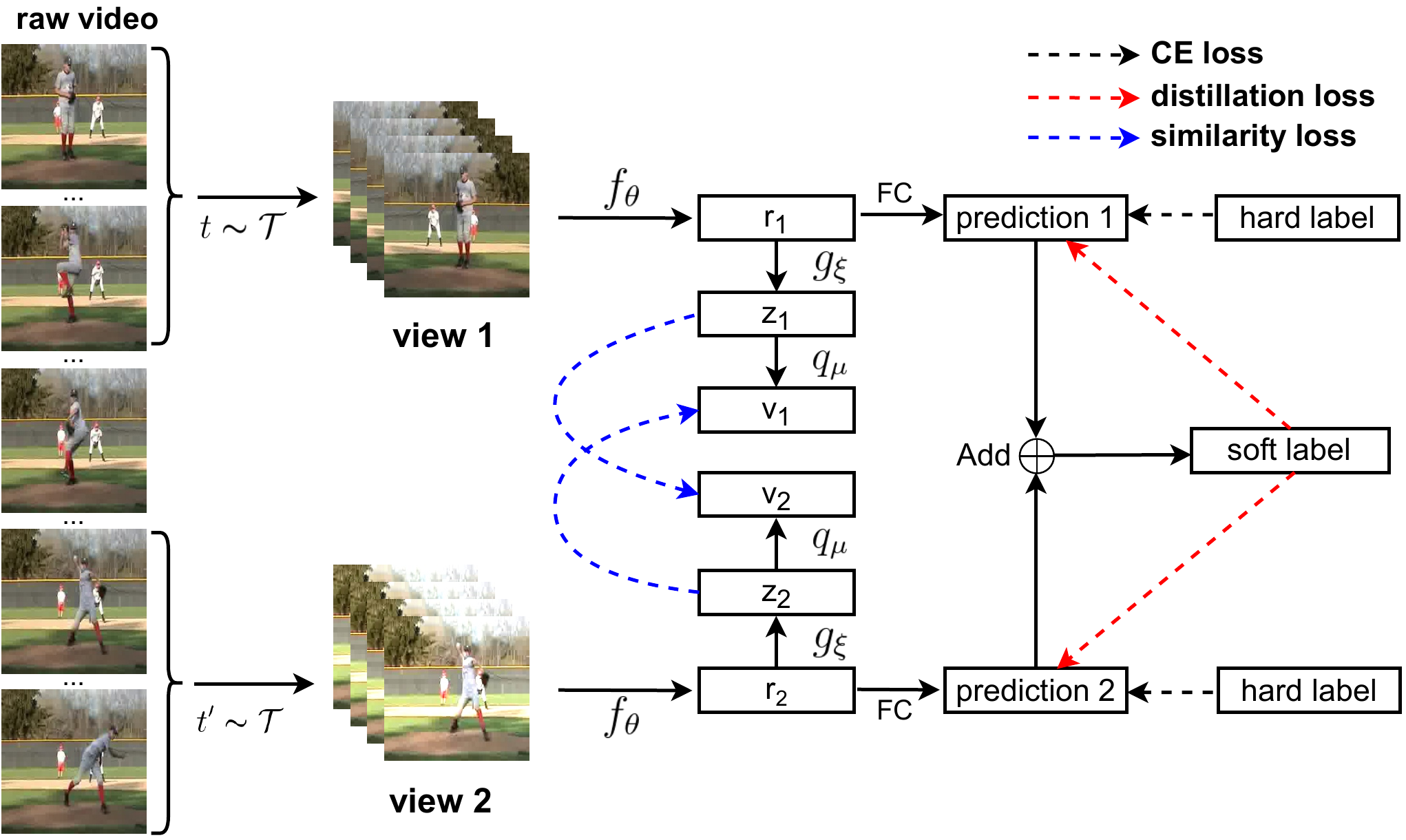}}
  \caption{\textbf{Overview of the SKD-SRL approach.} The black line is the forward path; CE loss is the standard cross-entropy loss; the distillation loss is calculated by Eq.~\ref{eq:loss_distill} and the similarity loss is calculated by Eq.~\ref{eq:loss_similarity}; $\oplus$ denotes the add operator. The hard label is the ground truth label of the video.}\medskip
  \label{fig:model}
\end{figure}

SKD-SRL is the combination of Self-KD and Siamese representation learning. We first apply data augmentation twice to obtain two versions (two views) of the input video. Each view is forward propagated through the encoder network $f_\theta$ to create a representation vector. Both representation vectors are passed into a fully connected ($FC$) layer to generate logit prediction vectors. The sum of logit predictions is utilized as the soft label to transfer the knowledge for each branch. Besides, both representation vectors are also forward to the projector MLP network $g_\xi$ and the predictor network $q_\mu$ to transform one vector's output and match it to the other vector by minimizing their negative cosine similarity. The proposed SKD-SRL method is illustrated in Figure.~\ref{fig:model}. In summary, our proposed approach includes three main steps as following:

\begin{enumerate}
    \item Calculating the representation vectors of two randomly augmented views from a given video by the encoder network $f_\theta$.
    
    \item Distilling the knowledge via soft labels, which is calculated through the representation vectors at step 1.
    
    \item Calculating the similarity between two representation vectors by Siamese representation learning.
\end{enumerate}

\subsection{Training Paradigm}
Given a set of $N$ training samples is denote as $\{(x^{(i)}, y^{(i)})\}_{i=1}^N$ as a labelled source dataset from $K$ classes where $x^{(i)}$ is a video, and $y^{(i)}$ is a $K$-dimensional one-hot vector as its hard label (i.e., the action). For a mini-batch input $B=\{(x^{(i)}, y^{(i)})\}_{i=1}^n$, we apply data augmentation operators (e.g. flip, contrast adjustment, etc.) for each $x^{(i)}$ in $B$. Let $x_1^{(i)}, x_2^{(i)} = \mathcal{T}(x^{(i)})$ denote the two randomly augmented views from the original video $x^{(i)}$ where $\mathcal{T}(\cdot)$ is the set of data augmentation operators. Note that $x_1^{(i)}, x_2^{(i)}$ have the same the label $y^{(i)}$. The two views $x_1^{(i)}, x_2^{(i)}$ are passed into an encoder network $f_\theta$, where $\theta$ denotes the set of parameters of $f$. Let $\mathbf{r}_1^{(i)}$ and $\mathbf{r}_2^{(i)}$ correspond to the output vectors of the network $f_\theta$ with the input being $x_1^{(i)}, x_2^{(i)}$ (i.e., $\mathbf{r}_1^{(i)} = f_\theta(x_1^{(i)})$ and $\mathbf{r}_2^{(i)} = f_\theta(x_2^{(i)})$). The encoder $f_\theta$ shares weights between the two views. A $FC$ layer is utilized to generate the logit prediction from $\mathbf{r}_1^{(i)}$ and $\mathbf{r}_2^{(i)}$ as follows:

\begin{equation}
\begin{split}
    \mathbf{p}_1^{(i)} = FC(\mathbf{r}_1^{(i)})\\
    \mathbf{p}_2^{(i)} = FC(\mathbf{r}_2^{(i)})
\end{split}
\label{eq:fc}
\end{equation}

where $\mathbf{p}_1^{(i)}$ and $\mathbf{p}_2^{(i)}$ denote the $K$-dimensional logit prediction vector and each dimension represents the logit value for the $k^{th}$ class (with $k=1,2,...,K$). Let $\mathbf{p}^{(i)} = \mathbf{p}_1^{(i)} \oplus \mathbf{p}_2^{(i)}$ where $\oplus$ corresponds to the add operator. Similar to other self-knowledge distillation methods, SKD-SRL also performs the knowledge distillation through the soft label as follows:


\begin{equation}
\begin{split}
    \mathcal{L}_{KL}(\mathbf{p}_1^{(i)}, \mathbf{p}_2^{(i)}; \tau) = D_{KL} \Big(softmax(\frac{\mathbf{p}^{(i)}}{\tau}) || softmax(\frac{\mathbf{p}^{(i)}_1}{\tau})\Big)  \\
    + D_{KL} \Big(softmax(\frac{\mathbf{p}^{(i)}}{\tau}) || softmax(\frac{\mathbf{p}_2^{(i)}}{\tau})\Big)
\end{split}
\label{eq:loss_distill}
\end{equation}

where $D_{KL}$ denotes the Kullback-Leibler (KL) divergence function, $\tau$ is the temperature scaling parameter. $\mathbf{r}_1^{(i)}$ and $\mathbf{r}_2^{(i)}$ also are passed into a projector MLP network $g_\xi$. We have $\mathbf{z}_1^{(i)} = g_\xi(\mathbf{r}_1^{(i)})$ and $\mathbf{z}_2^{(i)} = g_\xi(\mathbf{r}_2^{(i)})$. A predictor MLP $q_\mu$ is utilized to transform one vector $\mathbf{z}_1^{(i)}$ ($\mathbf{z}_2^{(i)}$) and match it to the other vector $\mathbf{z}_2^{(i)}$ ($\mathbf{z}_1^{(i)}$) by minimizing their negative cosine similarity, $\xi$ and $\mu$ denote the set of parameters of the network $g$ and $q$, respectively. Denoting the two output vectors as $\mathbf{v}_1^{(i)} = g_\mu(\mathbf{z}_1^{(i)})$ and $\mathbf{v}_2^{(i)} = g_\mu(\mathbf{z}_2^{(i)})$. We utilize the similarity loss function as follows:

\begin{equation}
\begin{split}
    \mathcal{L}_{sim}(\mathbf{z}_1^{(i)}, \mathbf{z}_2^{(i)},  \mathbf{v}_1^{(i)}, \mathbf{v}_2^{(i)}) = \frac{1}{2} D_{sim} (\mathbf{v}_1^{(i)}, stopgrad(\mathbf{z}_2^{(i)})) \\
    + \frac{1}{2} D_{sim} (\mathbf{v}_2^{(i)}, stopgrad(\mathbf{z}_1^{(i)}))
\end{split}
\label{eq:loss_similarity}
\end{equation}

where $stopgrad$ is the stop-gradient operator~\cite{chen2020exploring}. This means that $\mathbf{z}_1^{(i)}$ and $\mathbf{z}_2^{(i)}$ are treated as a constant in this term. $D_{sim}$ is the negative cosine similarity function as follows:

\begin{equation}
    D_{sim} (\mathbf{v}_1, \mathbf{z}_2) = - \frac{\mathbf{v}_1}{\left\lVert\mathbf{v}_1\right\rVert_2} \cdot \frac{\mathbf{z}_2}{\left\lVert\mathbf{z}_2\right\rVert_2},
\label{eq:d_sim}
\end{equation}

where $\left\lVert \cdot \right\rVert_2$ is $\ell_2$-norm. By integrating Eq.~\ref{eq:loss_distill} and Eq.~\ref{eq:loss_similarity}, we can construct the final optimization objective for the entire network as follows:

\begin{equation}
\begin{split}
    \mathcal{L}_{net} = \frac{1}{n} \sum_{i=1}^n \mathcal{L}_{CE}(y,softmax(\mathbf{p}_1^{(i)}))
    + \mathcal{L}_{CE}(y,softmax(\mathbf{p}_2^{(i)})) \\
    + \alpha \mathcal{L}_{KL}(\mathbf{p}_1^{(i)}, \mathbf{p}_2^{(i)}; \tau) + \beta \mathcal{L}_{sim}(\mathbf{z}_1^{(i)}, \mathbf{z}_2^{(i)},  \mathbf{v}_1^{(i)}, \mathbf{v}_2^{(i)})
\end{split}
\label{eq:net}
\end{equation}

where $\mathcal{L}_{CE}$ is the standard cross-entropy loss, $\alpha$ and $\beta$ are the loss weights for the distillation loss $\mathcal{L}_{KL}$ and the similarity loss $\mathcal{L}_{sim}$, respectively. The pseudo-code of SKD-SRL is illustrated in Algorithm.~\ref{alg:skd-srl}.

\begin{algorithm}
\DontPrintSemicolon
  \KwInput{$f_\theta$, $g_\xi$, $q_\mu$: the encoder, projector, and predictor networks, respectively. \;
  $\mathcal{T}$: the set of data augmentation operators \;
  $\alpha, \beta, \tau$: loss weights and temperature factor. \;
  }
  Initialize parameters $\theta, \xi, \mu$ \;
  \While{$\theta$ has not converged}
  {
    Sample a batch $(x, y)$ from the training set \;
    $x_1$, $x_2$ = $\mathcal{T}(x)$ \;
    $\mathbf{r}_1$, $\mathbf{r}_2$ = $f_\theta(x_1)$, $f_\theta(x_2)$ \;
    $\mathbf{p}_1$, $\mathbf{p}_2$  = $FC(\mathbf{r}_1)$, $FC(\mathbf{r}_2)$ \;
    $\mathbf{z}_1$, $\mathbf{z}_2$ = $g_\xi(\mathbf{r}_1)$, $g_\xi(\mathbf{r}_2)$ \;
    $\mathbf{v}_1$, $\mathbf{v}_2$ = $q_\mu(\mathbf{z}_1)$, $q_\mu(\mathbf{z}_2)$ \;
    Calculate the loss $\mathcal{L}_{net}$ by Eq.~\ref{eq:net}\;
    Update parameters $\theta$, $\xi$, $\mu$ 
  }
  \Return Encoder network $f_\theta$
\caption{SKD-SRL Pseudocode for Action Recognition}
\label{alg:skd-srl}
\end{algorithm}

\subsection{Network Architecture and Data Augmentation}
\textbf{Network Architecture.} For the encoder network $f_\theta$, we consider two state-of-the-art CNN architectures including the 3D ResNet-18, and the 3D ResNet-50 in~\cite{resnet3D_50}. For the projector network $g_\xi$ which maps the representation vectors $\mathbf{r}_1$ and $\mathbf{r}_2$ to vectors $\mathbf{z}_1$ and $\mathbf{z}_2$, we instantiate $g_\xi$ just a single linear layer of size = 2048. The predictor network $q_\mu$ has two MLP layers where the first hidden layer is applied BN and ReLU, and the output layer does not have BN and activation function. Due to the predictor's input and output dimension as 2048, we set the predictor's hidden layer’s dimension as 512 following the instruction in~\cite{chen2020exploring}. We leave to future work the investigation of optimal $g_\xi$ and $q_\mu$ architectures.

\textbf{Data Augmentation.} Given an input video, our data augmentation method takes two steps. We first trim two clips with $T$ continuous frames from the raw video. Each frame in the clip is scaled the shorter edge of the frames in the clip to 128, and the other edge is calculated to maintain the frame original aspect ratio. A random cropping window with dimensions of $112 \times 112$ is generated and applied to all frames. We then randomly choose data augmentation operators to apply for each clip to generate the two views from the original clip. The list of all augmentation operators are used in our work including flip, contrast adjustment, brightness adjustment, hue adjustment, Gaussian blur and channel splitting~\cite{vu2021teaching}. The probability chosen for each augmentation operator is set to 0.5.

\section{Experiment}
\label{sec:experiment}
\subsection{Datasets and Implementation}
\noindent
We have conducted experiments on three datasets including UCF101~\cite{ucf101}, HMDB51~\cite{hmdb51} and Kinetics400~\cite{kinetics}.

\textbf{UCF101:} includes 13,320 action instances from 101 human action classes. The average duration of each video is about 7 seconds. 

\textbf{HMDB51:} is a small dataset including 6,766 videos from 51 human action classes. The average duration of each video is about 3 seconds. 

\textbf{Kinetics400:} is a large dataset that has 400 human action classes~\cite{kinetics}. The videos were temporally trimmed and last around 10 seconds and 200–1000 clips for each action. The total has 306,245 videos in Kinetics400. 

\textbf{Implementation Details}. All networks are trained from scratch and optimized by stochastic gradient descent (SGD) with a momentum of 0.9 and an initial learning rate of 0.01. The weight decay is set to $5 \times 10^{-4}$. The input of the encoder network is a video clip with 16 frames, each frame has $112 \times 112 \times 3$ of dimension and normalized to be [-1, 1]. We use the mini-batch of 32 clips per GPU, and training is done in 200 epochs. The learning rate is dropped by 10$\times$ after 10 epochs if the validation accuracy not improving. For our method, the temperature $\tau$ is set as 10, and the loss weights $\alpha$ and $\beta$ are set as 0.1 and 1, respectively.

\subsection{Comparison with independently training}
To examine the effectiveness of the proposed SKD-SRL method, we have compared our approach to the baseline (independently training) with cross-entropy loss on the standard datasets with both the ResNet-18 and ResNet-50 networks.

\begin{table}[!ht]
    \centering
    \setlength\tabcolsep{5pt}
    \renewcommand{\arraystretch}{1.3}
    \caption{Top-1 accuracy of the SKD-SRL compares to the baseline method in both the 3D ResNet-18 and the 3D ResNet-50 networks on standard datasets.}
    \begin{tabular}{lcccc}
    \hline \hline
    \textbf{Method}  & \textbf{Backbone}     & \textbf{UCF101} & \textbf{HMDB51} & \textbf{Kinetics400} \\
    \hline \hline
    Baseline & ResNet-18 & 46.5 & 17.1 & 54.2\\
    \textbf{SKD-SRL}& ResNet-18 & \textbf{69.8} & \textbf{24.7} & \textbf{66.7} \\ \hline
    Baseline & ResNet-50 & 59.2 & 22.0 & 61.3\\
    \textbf{SKD-SRL} & ResNet-50 & \textbf{71.9} & \textbf{29.8} & \textbf{75.6} \\
    \hline 
    \hline
    \end{tabular}
 \label{tab:compare_to_random_init}
\end{table}

As shown in Table.~\ref{tab:compare_to_random_init}, the SKD-SRL method outperforms the baseline method for both the large and small-scale datasets regardless of the backbone networks. Specifically, the SKD-SRL method increases the accuracies by 23.3\% and 12.7\% for the 3D ResNet-18 and 3D ResNet-50 networks, respectively, on small-scale datasets such as the UCF101 dataset. In the large-scale dataset, i.e., the Kinetics400 dataset, the SKD-SRL method increases the accuracies by 12.5\% and 14.3\% for the 3D ResNet-18 and 3D ResNet-50 networks, respectively. From these results, we found that our approach can significantly improve generalization and performance compared to independently training only with hard labels.

\subsection{Comparison with other KD mechanisms}
Table~\ref{tab:compare_to_distillation} compares our SKD-SRL method with other distillation mechanisms. As expected, the student performance in distillation approaches improves compared to independent training (i.e., baseline). Moreover, the Self-KD method shows better top-1 accuracy than the conventional KD (+5.3\% accuracy). This shows that although there is no teacher network, Self-KD still significantly improves performance via its self-teaching and self-learning mechanism. Meanwhile, the proposed SKD-SRL method outperforms 3.1\% compare to Self-KD. It demonstrates that our approach enhances the generalization capability of the single network by combining Self-KD with Siamese representation learning.

\begin{table}[!ht]
    \centering
    \setlength\tabcolsep{10pt}
    \renewcommand{\arraystretch}{1.3}
    \caption{Comparison with different distillation mechanisms on the UCF101 dataset. The student network in all mechanisms is the ResNet-18 network.}
    \begin{tabular}{lccc}
    \hline \hline
    \textbf{Mechanism}  &  \textbf{Teacher Network} & \textbf{Top-1 Accuracy} \\
    \hline \hline
    
    Baseline  & None  & 46.5 \\ 
    Baseline + Data augment  & None  & 55.6 \\ 
    
    KD &  ResNet-50  & 61.4 \\ 
    
    Self-KD & itself   & 66.7 \\ 
    
    \textbf{SKD-SRL} & itself & \textbf{69.8}\\
    \hline \hline
    \end{tabular}
 \label{tab:compare_to_distillation}
\end{table}

\subsection{Comparison with state-of-the-art methods}
In this part, we evaluate SKD-SRL on the Kinetics400 dataset with two backbone networks, including 3D ResNet18 and 3D ResNet-50. As shown in Table.~\ref{tab:result_compare_supervised}, the proposed SKD-SRL method has obtained state-of-the-art performance while utilizing fewer frames (16 vs 32, 64) and lower frame resolutions (112 vs 224). In particular, the SKD-SRL method uses RGB frames without the optical flow, which significantly reduces the cost of optical flow calculation and model training on the optical flow domain. Moreover, our proposed SKD-SRL has better performance with a shallower model than other state-of-the-art methods (3D ResNet-50 vs 3D ResNet-101, DenseNet-169).

\begin{table}[!ht]
  \centering
  \medskip
  \setlength\tabcolsep{3pt}
  \renewcommand{\arraystretch}{1.3}
  \caption{Top-1 accuracy of the SKD-SRL method compares to state-of-the-art methods on the Kinetics400 dataset. * indicates that these methods use both RGB and optical flow frames in the training phase.}
\begin{tabular}{lcccccc}
\hline \hline
\textbf{Method}   & \textbf{Backbone}   & \begin{tabular}[c]{@{}c@{}} \textbf{Pretraining} \\ \textbf{dataset}  \end{tabular} & \textbf{\#frames} & \begin{tabular}[c]{@{}c@{}} \textbf{Frame} \\ \textbf{resolution} \end{tabular} & \begin{tabular}[c]{@{}c@{}} \textbf{Top-1} \\  \textbf{Accuracy} \end{tabular}  \\ 
\hline 
\hline

I3D~\cite{i3d_2017} & InceptionNet & ImageNet & 64 & $224 \times 224$ & 71.1 \\

FASTER~\cite{zhu2020faster} & R(2+1)D-50 & None & 32 & $224 \times 224$ & 71.7 \\


R(2+1)D*~\cite{tran2018closer} & ResNet-34 & Sport-1M & 32 & $112 \times 112$ & 73.3 \\

bLVNet~\cite{NEURIPS2019_3d779cae} & ResNet-50 & None & 24 & $224 \times 224$ & 73.5 \\

MoViNets~\cite{kondratyuk2021movinets} & MoViNet-A1 & None & 50 & $224 \times 224$ & 72.7 \\

MoViNets~\cite{kondratyuk2021movinets} & MoViNet-A2 & None & 50 & $224 \times 224$ & 75.0 \\

MViT~\cite{fan2021multiscale} & MViT-S & None & 16 & $224 \times 224$ & \textbf{76.0} \\


R3D~\cite{resnet3D_50} & ResNet-152 & None  & 16 & $112 \times 112$ & 63.0 \\

R3D~\cite{resnet3D_50} & ResNeXt-101 & None & 16 & $112 \times 112$ & 65.1 \\

STC~\cite{diba2018spatio} & ResNeXt-101 & ImageNet  & 32 & $112 \times 112$ & 68.7 \\

T3D~\cite{t3d} & DenseNet-169 & ImageNet  & 32 & $224 \times 224$ & 62.2 \\

MARS*~\cite{crasto2019mars} & ResNeXt-101 & None  & 16 & $112 \times 112$ & 68.9 \\


\hline 

\textbf{SKD-SRL} & ResNet-18 & None & 16 & $112 \times 112$ & 66.7 \\
\textbf{SKD-SRL} & ResNet-50 & None  & 16 & $112 \times 112$ & \textbf{75.6} \\

\hline 
\hline
\end{tabular}

\label{tab:result_compare_supervised}
\end{table}

\section{Conclusion}
In this work, we have introduced the combination of Self-knowledge distillation and Siamese representation learning. Our approach utilizes the sum of predictive distributions from two different views of the same sample for the soft label distillation. Moreover, we minimize the difference between two representation vectors of these views via Siamese representation learning. Experiments conducted across different network architectures have shown that our proposed method achieves state-of-the-art performance compared to other methods on the standard datasets. In future work, we investigate optimal architectures for projector and predictor networks. In addition, the proposed method can be adapted and applied to other computer vision tasks involving.





%
\bibliographystyle{IEEEbib}
\bibliography{refs}
\end{document}